\documentclass[3p,twocolumn]{elsarticle}

\usepackage{lineno,hyperref}
\modulolinenumbers[5]

\usepackage{graphicx}
\usepackage{amsmath}
\usepackage{amssymb}
\usepackage{amsfonts}
\usepackage{multirow}
\usepackage{rotating}
\usepackage{subfig}
\usepackage{color}
\usepackage{natbib}
\usepackage{float}
\usepackage{url}

\definecolor{myred}{rgb}{.8,.0,.0}

\begin{document}

\begin{frontmatter}

\title{Not-so-supervised: a survey of semi-supervised, multi-instance, and transfer learning in medical image analysis}

\address[bigr]{Biomedical Imaging Group Rotterdam, Depts. Radiology and Medical Informatics, Erasmus Medical Center, Rotterdam, The Netherlands}
\address[image]{Medical Image Analysis, Dept. Biomedical Engineering, Eindhoven University of Technology, Eindhoven, The Netherlands}
\address[isi]{Image Sciences Institute, University Medical Center Utrecht, Utrecht, The Netherlands}
\address[diku]{The Image Section, Dept. Computer Science, University of Copenhagen, Copenhagen, Denmark}

\author[image]{Veronika Cheplygina\corref{cor1}}
\ead{v.cheplygina@tue.nl}
\cortext[cor1]{Corresponding author}

\author[bigr,diku]{Marleen de Bruijne}

\author[image,isi]{Josien P. W. Pluim}

\begin{abstract}
Machine learning (ML) algorithms have made a tremendous impact in the field of medical imaging. While medical imaging datasets have been growing in size, a challenge for supervised ML algorithms that is frequently mentioned is the lack of annotated data. As a result, various methods that can learn with less/other types of supervision, have been proposed. We give an overview of semi-supervised, multiple instance, and transfer learning in medical imaging, both in diagnosis or segmentation tasks. We also discuss connections between these learning scenarios, and opportunities for future research.  \\
\end{abstract}

\end{frontmatter}

\section{Introduction}\label{sec:introduction}

Machine learning has become very important in medical image analysis. Tasks such as segmentation, where each pixel or voxel in an image is assigned to a different anatomical structure or tissue type, and  computer-aided diagnosis where a category label or a continuous value is predicted for an entire image, are now almost exclusively done with machine learning methods.

A frequent problem when applying machine learning methods to medical images, is the lack of labeled data~\citep{litjens2017survey,weese2016four,bruijne2016machine}, even when larger sets of unlabeled data may be more widely available. An important reason for this is the sheer difficulty of collecting the labels. Manual labeling of the images is an expensive and/or time-consuming process. Such labels might not be needed in clinical practice, therefore reducing the availability of labeled data only to research studies. Another issue is that, even when labeled data is collected, it is not often available to other researchers.

The lack of labeled data motivates approaches that go beyond traditional supervised learning by incorporating other data and/or labels that might be available. These approaches include semi-supervised learning, multiple instance learning and transfer learning, although many other terms exist to describe these approaches. Papers within one of these learning scenarios seem to be aware of other related literature, and surveys are emerging, such as \citep{quellec2017multiple}. However, it seems that there is little interaction between the scenarios, which is a missed opportunity, since their goals are related.

With this survey, we aim to provide an overview of the learning scenarios, describe their connections, identify gaps in the current approaches, and provide several opportunities for future research.  The survey is primarily aimed at researchers in medical image analysis. We have however made an effort for the survey to also be accessible to a broader readership.

\subsection{Selection of papers}

An initial selection of papers was created by screening the results of Google Scholar searches for terms ``semi-supervised learning'', ``multiple instance learning'' and ``transfer learning'' for medical imaging papers. These papers were used to identify other relevant publications. In the event of multiple similar papers, only the latest paper was included. Only papers that became available online before 2018 were included. After publishing the preprint online, we received more suggestions for relevant papers, and included these if these fit our criteria.

This survey does not cover approaches that rely on interaction with the annotator, such as active learning or crowdsourcing, in detail. We focus on machine learning approaches that can be used even if there is no possibility of acquiring additional labels. We focus on classification tasks within medical image analysis, for diagnosis, detection or segmentation purposes.

It is our intention to provide an overview how different learning scenarios are being used rather than a full summary of all related papers. We emphasize that we focus on the types of learning scenarios and the assumptions that are being made, rather than specific classifiers. 

\section{Overview of techniques}

In this section we provide a quick overview of the learning scenarios. We also provide examples of each type of learning scenario, based on the application of classifying emphysema, a sign of chronic obstructive pulmonary disease (COPD) in chest computed tomography (CT) images. For readability, we provide a list of notation and acronyms that will be introduced throughout the paper in Tables~\ref{tab:notation} and~\ref{tab:acronyms}.

\begin{figure*}
\centering
    \includegraphics[width=\textwidth]{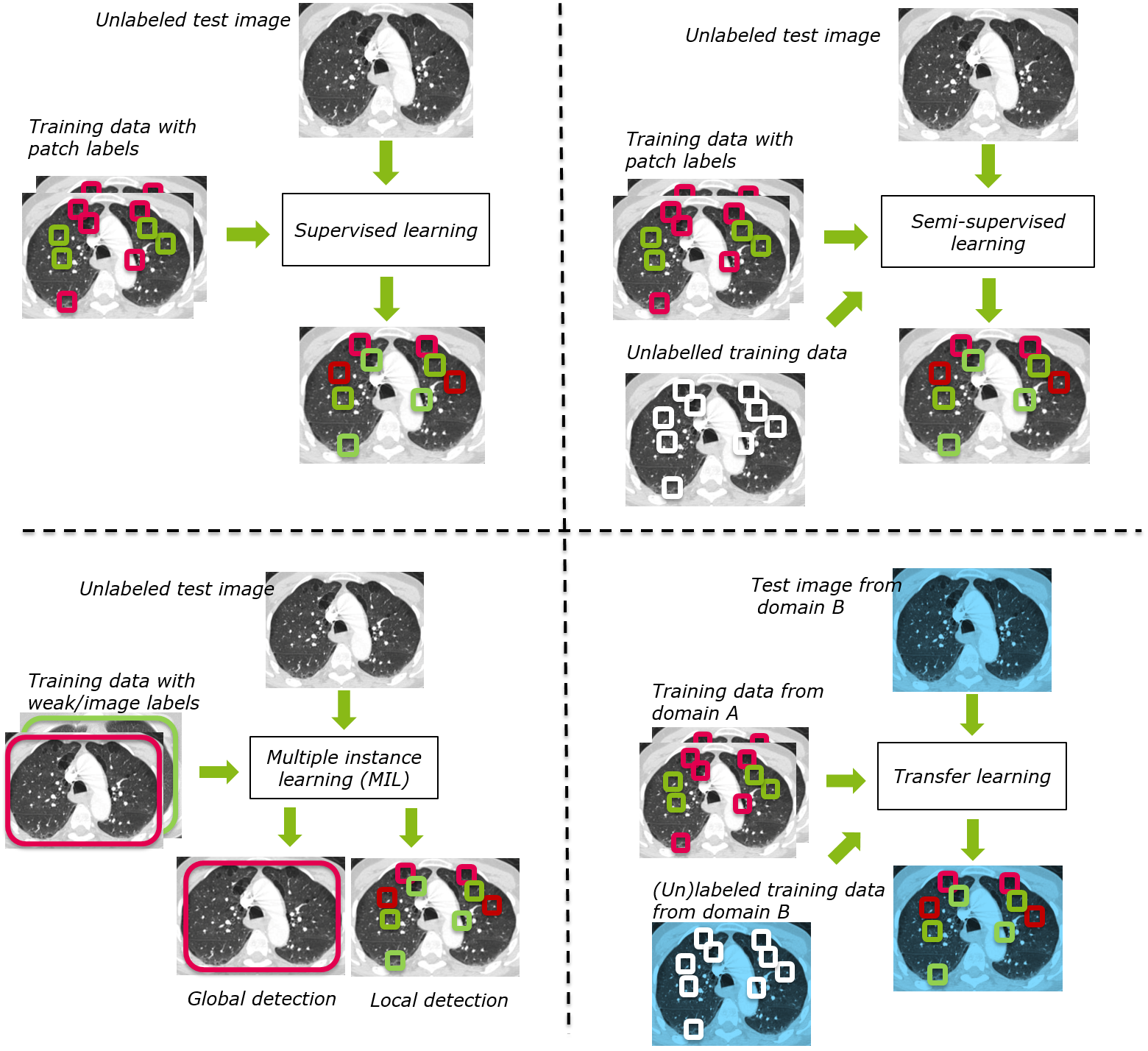}
    \caption{Learning scenarios, illustrated by a task of classifying healthy (green) vs emphysema (red) tissue in chest CT images. Annotations are made for presentation purposes only and do not necessarily reflect ground truth. \textbf{Top left:} Supervised learning, only healthy and abnormal patches are available. \textbf{Top right:} Semi-supervised learning (Section \ref{sec:ssl}). In addition to healthy and abnormal patches, unlabeled patches are available. \textbf{Bottom left:} Multiple instance learning (Section \ref{sec:mil}). Labeled patches are not available, but subject-level labels (whether any abnormal patches are present) are. \textbf{Bottom right:} Transfer learning (Section \ref{sec:tl}). Labeled patches are available, but for a different domain (here illustrated by different visual characteristics) than in the test set.}
    \label{fig:learning_scenarios}
\end{figure*}

In supervised learning, we have a training set $D_S = \{(\mathbf{x}_i\, y_i\}$, where each sample $\mathbf{x}_i \in \mathcal{X}$ is associated with a label $y_i \in \mathcal{Y}$. Here $\mathcal{X}$ is the feature space, such as an $m$-dimensional space of real numbers $\mathbb{R}^m$, and $\mathcal{Y}$ is the label space, such as the set $\{0,1\}$ for a binary classification problem or the set $\mathbb{R}$ for a regression problem. We want to use this data to train a classifier $f: \mathcal{X} \rightarrow \mathcal{Y}$ that can provide labels for unlabeled samples from a previously unseen test set $S_T$. For example, the instances are patches in a chest CT image, and they are labeled as emphysema or as normal tissue. At test time, we want to classify all patches in a previously unseen scan as emphysema or not. This example is illustrated in Fig.~\ref{fig:learning_scenarios}(a). 

In \emph{semi-supervised learning} (SSL), in addition to the training set we have an unlabeled set of data $U$. We want to use this set to improve the predictions of the classifier on $D_T$. For example, the supervised problem above can be extended with patches from chest CT images that have not been manually labeled by experts. This scenario is covered in Section~\ref{sec:ssl} and illustrated in Fig.~\ref{fig:learning_scenarios}(b).

In \emph{multiple instance learning} (MIL), the training set itself consists of \emph{bags} of instances $X_i = \{\mathbf{x}_{ij}, j=1,\ldots,N_i\}$. The bags are labeled. The instances have unknown labels $y_{ij}$ that are somehow related to the bag label $Y_i$. An example of such a relationship is ``if at least one instance is positive, the bag is also positive''. For example, this situation can occur if the radiologist only labeled an entire CT scan as containing emphysema or not, but has not indicated its locations.

In MIL the test set also consists of bags, which are unlabeled. Next the goal can be two-fold: to classify the test bags and/or to classify the test instances. In our example, the bag classifier would predict whether a patient has any emphysema, whereas the instance classifier would in addition localize any emphysema in the image.  This scenario is covered in Section~\ref{sec:mil} and illustrated in Fig.~\ref{fig:learning_scenarios}(c).

In the scenarios above, we assume that the training and test data are from the same \emph{domain} $\mathcal{D} = (\mathcal{X}, p(\mathbf{x})$), defined by the feature space and distribution of the samples. However, this is not always the case, creating a \emph{transfer learning} (TL) scenario. In this scenario we assume to have a source dataset $D_S$ where the instances $\mathbf{x}_{S_i} \in \mathcal{X}_S$ and a test or target set $D_T$ where  the instances $\mathbf{x}_{T_i} \in \mathcal{X}_S$. For example, this can occur when different scanning protocols are used, leading to different appearance of the patches, and therefore different distributions $p(\mathbf{x}_S)$ and $p(\mathbf{x}_T)$. The goal is to train a classifier using $D_S$, and possibly using either the unlabeled test data $D_T$, and/or labeled data from the target domain $L \in \mathcal{D}_T$. This scenario is covered in Section~\ref{sec:tl} and illustrated in Fig.~\ref{fig:learning_scenarios}(d). As we discuss later, this scenario is not limited to the case where there are differences in the feature distributions.

In Section~\ref{sec:discussion} we discuss the trends within these learning scenarios, the gaps in the current research, and the opportunities and challenges for future research.

\begin{table}[]
    \centering
    \begin{tabular}{l|l}
         Feature space & $\mathcal{X}$ \\
         Label space & $\mathcal{Y}$ \\
         Classifier & $f: \mathcal{X} \rightarrow \mathcal{Y}$ \\
         
         Instance &  $\mathbf{x}_i \in \mathcal{X}$ \\
         Instance label & $y_i \in \mathcal{Y}$ \\
         Bag & $X_i = \{\mathbf{x}_{ij}, j=1,\ldots,N_i\}$ \\
         Bag label & $Y_i \in \mathcal{Y}$\\
         
         Domain & $\{\mathcal{X}, p(\mathbf{x})\}$\\
         Domain (MIL) & $\{\mathcal{X}, p(X)\}$\\
         Task & $\{\mathcal{Y}, f(\cdot) \}$\\
     
         Subscript $_S$ & source \\
         Subscript $_T$ & target \\
         Source domain & $\mathcal{D}_S$ \\
         Target domain & $\mathcal{D}_T$ \\
         Source task & $\mathcal{T}_S$ \\
         Target task & $\mathcal{T}_T$ \\
         
         Training (source) data & $D_S$ \\
         Test (target) data & $D_T$ \\
         Unlabeled (source) data & $U$ \\
         Labeled (target) data & $L$ \\

    \end{tabular}
    \caption{Notation used throughout the paper}
    \label{tab:notation}
\end{table}

\begin{table}[]
    \centering
    \begin{tabular}{l|l}
    ML & machine learning \\
    SSL & semi-supervised learning \\
    MIL & multiple instance learning \\
    TL & transfer learning \\
    MTL & multi-task learning \\
    SVM & support vector machine \\

    \hline
    AD & Alzheimer's disease \\
    MCI & mild cognitive impairment \\
    COPD & chronic obstructive pulmonary disease \\
    CT & computed tomography \\
    DR & diabetic retinopathy \\
    MR & magnetic resonance \\
    US & ultrasound \\
    \end{tabular}
    \caption{Acronyms used throughout the paper}
    \label{tab:acronyms}
\end{table}

\section{Semi-supervised learning}\label{sec:ssl}

In the semi-supervised learning scenario, there are two sets of samples: labeled samples $D_S$ and unlabeled samples $U$. The goal is to use the samples in $U$ to improve the classifier $f$, where $f$ is constructed only using samples in $D_S$. For example, when classifying emphysema vs normal patches, the scans that have been annotated are used to create a set of labeled patches, while the scans without annotations can be used to create a large unlabeled set of patches. We can distinguish two goals in SSL: predicting labels for future data (inductive SSL) and predicting labels for the already available unlabeled samples (transductive SSL)~\citep{zhu2009introduction}.

Typically semi-supervised approaches work by making additional assumptions that link properties of the distribution of the input features to properties of the decision function ~\citep{chapelle2006semi,zhu2009introduction}. These include the \emph{smoothness assumption}, i.e. samples close together in feature space are likely to be from the same class, the \emph{cluster assumption}, i.e. samples in a cluster are likely to be from the same class, and the \emph{low density assumption}, i.e.  class boundaries are likely to be in areas of the feature space that have lower density than the clusters. 

Many semi-supervised approaches therefore proceed with exploiting such assumptions. A popular method called self-training propagates labels from the labeled to the unlabeled data, and then using the larger, newly labeled set for training. This approach assumes that the method's high confidence predictions are correct, which is likely to be the case with the cluster assumption. Expectation-maximization uses a principle similar to self-training by alternating between assigning soft labels to the unlabeled data given the labeled data and model parameters, and updating the model parameters given all the data. Another related approach is co-training, where classifiers are trained with independent sets of features, and the classifiers rely on each other for estimating the confidence of their predictions. 

Other popular methods include methods that regularize the classifier based on the unlabeled data, such as graph-based methods and semi-supervised support vector machines (SVMs). Graph-based methods encode similar samples as connected nodes and solve a graph cut problem, therefore assuming low density between classes. Semi-supervised SVMs encourage margins that place unlabeled data outside the margin, also assuming low density separation. An overview of methods and corresponding assumptions can be found in \cite{zhu2009introduction}. 

When the additional assumptions do not hold, there is a risk of performing worse than a supervised approach~\citep{cozman2006risks,zhu2009introduction}. More recent are approaches that do not make additional assumptions about the data, and instead use assumptions already present in the classifier~\citep{loog2015semi,krijthe2017robust}. For example, this can be achieved by linking parameter estimates (such as mean and variance of the samples) based on labeled samples, to those based on all available samples.

\subsection{SSL in medical imaging}

SSL is a naturally occurring scenario in medical imaging, both in segmentation and diagnosis tasks. In segmentation methods, an expert might label only a part of the image, leaving many samples unlabeled. In computer-aided diagnosis, there might be ambiguity about the label of a subject, so rather than adding these subjects to the training set or removing them completely, they can still be used without labels to improve the classifier. For example, in classification subjects as having Alzheimer's disease (AD) or normal cognitive function (CN), subjects with mild cognitive impairment (MCI) who may or may not develop AD later, are sometimes considered unlabeled~\citep{filipovych2011semi,batmanghelich2011disease}.

The papers using SSL are summarized in Table \ref{tab:ssl}. Overall there are two main directions. In the first, papers use a self-training or co-training approach for segmentation purposes. We discuss these in Section \ref{sec:ssl_selftraining}. In the second, papers use the unlabeled data to regularize the classifier via graph-based methods or SVMs. These approaches are used both for segmentation and diagnosis tasks. We discuss these papers in Section \ref{sec:ssl_regularization}.

\begin{table*}[]
\centering
\scriptsize
\begin{tabular}{l p{5cm} l}

Reference & Application & SSL category\\

\hline
\multicolumn{3}{c}{Brain}\\
\hline

\cite{song2009semi} & tumor segmentation & graph-based\\
\cite{iglesias2010agreement} & skull stripping & self-training\\ 
\cite{filipovych2011semi} & classification of MCI & semi-supervised SVM\\
\cite{batmanghelich2011disease} & classification of AD, MCI & graph-based\\
\cite{xie2013multiple} & tissue segmentation & graph-based \\

\cite{meier2014patient} & tumor segmentation & graph-based\\ 
\cite{dittrich2014spatio} & fetal brain segmentation & self-training \\
\cite{wang20144d} & lesion segmentation & self-training, active \\
\cite{an2016semi} & AD classification & graph-based \\
\cite{baur2017semi} & MS lesion segmentation & graph-based \\
\cite{moradi2015machine} & classification of MCI & semi-supervised SVM\\

\hline
\multicolumn{3}{c}{Eye}\\
\hline

\cite{adal2014automated} & microaneurysm detection & self-training\\
\cite{mahapatra2016combining} & optic disc missing annotation prediction & self-training, graph-based \\

\hline
\multicolumn{3}{c}{Breast}\\
\hline
\cite{sun2016computerized} & mass classification & co-training \\

\hline
\multicolumn{3}{c}{Heart}\\
\hline
\cite{zuluaga2011learning} & detection of vascular lesions  & self-training \\ 
\cite{bai2017semi} & cardiac segmentation & self-training \\
\cite{wang2017direct} & aneurysm volume estimation & graph-based \\

\hline
\multicolumn{3}{c}{Lung}\\
\hline
\cite{prasad2009multi} & segmentation of emphysema in CT & self-training / co-training, active\\
\cite{rikxoort2010multi} & classification of tuberculosis patterns in CT & self-training / co-training \\

\hline
\multicolumn{3}{c}{Abdomen}\\
\hline
\cite{tiwari2010semi} & classification of cancerous areas in prostate & graph-based \\
\cite{park2014interactive} & prostate segmentation  & graph-based, active \\
\cite{borga2016semi} & liver segmentation & graph-based \\ 
\cite{mahapatra2016combining} & predicting missing expert annotations of Crohn's disease & self-training, graph-based \\ 

\hline
\multicolumn{3}{c}{Histology and microscopy}\\
\hline
\cite{singh2011identifying} & cell type classification in microscopy & self-training \\
\cite{parag2014small} & cell type segmentation in microscopy & graph-based, active \\
\cite{xu2016neuron} & neuron segmentation in microscopy & graph-based \\
\cite{su2016interactive} & cell segmentation in microscopy & graph-based, active \\

\hline
\multicolumn{3}{c}{Multiple}\\
\hline
\cite{gass2013semi} & segmentation in two applications & graph-based \\
\cite{ciurte2014semi} & segmentation in four applications & graph-based\\
\cite{gu2017semi} & segmentation in two applications & self-training \\

\hline
\multicolumn{3}{c}{Other}\\
\hline
\cite{huang2008semi} & segmentation of nasopharyngeal
carcinoma lesion in MR & graph-based \\

\end{tabular}
\caption{Overview of semi-supervised learning applications. The last column describes the type of method used, ``active'' refers to active learning.}
\label{tab:ssl}
\end{table*}

\normalsize

\subsection{Self-training and co-training}\label{sec:ssl_selftraining}

A popular approach to SSL in medical imaging is label propagation via self-training. The general idea is as follows. A classifier is first trained on the labeled samples. The trained classifier then classifies the unlabeled samples. These samples, or a subset of these samples, are then added to the training set.  This process is repeated several times.

The surveyed papers differ in how they select the subset of unlabeled samples to add to the labeled data. Several papers choose an active learning approach, where expert interaction is needed to verify some of the labels~\citep{parag2014small,su2016interactive}. As mentioned in the introduction, we do not in detail address active learning, and only focus on methods that can be used even if no additional labels can be acquired.

Other papers measure the uncertainty or confidence of the classifier based on the output (posterior probability) of the classifier itself, and possibly additional classifiers.  \citet{wang20144d} add samples with a confidence above a user-selected threshold to the training set. Additional classifiers can be used as well, in which case the method falls under co-training. For example, for skull segmentation, \cite{iglesias2010agreement} use two conventional tools and their own classifier, to classify all the unlabeled pixels. The pixels for which the conventional tools agree, but their own classifier is not confident, are added to the labeled set. A similar strategy is used by \citet{rikxoort2010multi} for classifying tuberculosis patterns in chest CT, but with simple classifiers like $k$ nearest neighbor to estimate agreement.

Self-training is popular for segmentation, for propagating labels between pixels/voxels. It is used in the brain~\citep{iglesias2010agreement,meier2014patient,wang20144d,dittrich2014spatio}, retina~\citep{gu2017semi}, heart~\citep{bai2017semi} and several other applications. Self-training is less common for computer-aided detection or diagnosis applications. In the surveyed papers,  \citep{rikxoort2010multi} classify volumes of interest in chest CT and  \citep{singh2011identifying} classify cell nuclei, but in both cases the sample size is in the thousands. This suggests that self-training is more often used for applications with larger datasets, which are common in segmentation but less so in computer-aided diagnosis.

A few works investigate how the sample size affects performance. \citet{iglesias2010agreement,bai2017semi,gu2017semi} all  show that increasing the number of samples increases performance, and that the advantages of semi-supervised methods decrease as more labeled data becomes available.

\subsection{Graph-based methods and regularization} \label{sec:ssl_regularization}

Another popular strategy is to use the unlabeled data to better estimate the distribution of the data, and as such, regularize the classifier. Graph-based methods and semi-supervised SVMs fall under this category, but make different assumptions about the data. 

Graph-based methods construct a graph with the samples as nodes, and similarities of these samples (defined via a distance measure and/or prior knowledge) as edges. The assumption is that connected samples are likely to have the same label, and the goal is to propagate the labels along the graph. This can be achieved with a graph cut algorithm, which finds a labeling of the samples such that the outputs for the labeled training samples are correct, and the outputs of all samples are smooth along the graph. However, finding a labeling means that previously unseen images cannot be labeled without running the procedure again, also referred to as the out-of-sample problem. In the surveyed papers, graph cuts are often used for segmentation~\citep{mahapatra2016active,song2009semi,ciurte2014semi,wang20144d,su2016interactive}, the labels are therefore propagated between pixels or superpixels. This means that for a previously unseen image that needs to be segmented, some labeled pixels would be needed. 

Graph-based methods can be also used for atlas-based segmentation~\citep{gass2013semi,borga2016semi}, but with an important difference. Instead of constructing a graph of pixels, atlas-based segmentation methods construct a graph of images. When segmenting a test image, the labels of a single atlas are first propagated to the (unlabeled) images that are neighbors on this graph. These atlases with propagated labels can then be combined into a final labeling.

Manifold regularization uses a similar idea of smoothness along a graph, and is able to label previously unseen data. Here the graph Laplacian encodes the similarity of the nodes and is used as a regularizer, encouraging smoothness along the graph. This method is used both for segmentation~\citep{song2009semi,park2014interactive,xu2016neuron} and computer-aided diagnosis~\citep{tiwari2010semi,an2016semi,batmanghelich2011disease,wang2017direct}.

Semi-supervised SVMs use a different assumption, namely that there is a low density region between the classes. Next to fitting a hyperplane using the labeled training samples, semi-supervised SVMs also try to enforce this assumption, by favoring hyperplanes that place unlabeled samples outside the margin. This approach is used for classification of AD or MCI~\citep{filipovych2011semi,moradi2015machine}.

\section{Multiple instance learning}\label{sec:mil}

The multiple-instance learning (MIL) scenario can occur when obtaining ground-truth local annotations (i.e. for pixels or patches) is costly, time-consuming or not possible, but global labels for whole images, such as the overall condition of the patient, are available more readily. However, these labels do not apply to all the pixels or patches inside the image. MIL is an extension of supervised learning that can train classifiers using such weakly labeled data. For example, a classifier trained on images (\emph{bags}), where each bag is labeled as normal or abnormal and consists of unlabeled image patches (\emph{instances}), would be able to label novel images, and/or patches of that image as normal or abnormal.

A sample is a \emph{bag} or set $X_i  = \{\mathbf{x}_{ij}| j=1,...,N_i\} \subset \mathbb{R}^m$ of $N_i$ instances, each instance is thus a $m$-dimensional feature vector. We are given labeled training bags $\{(X_i, Y_i) | i=1,...N_{S}\}$ where $Y_i$ is the label. Originally MIL was formulated as a binary classification problem, but multi-class generalizations have also been proposed. For simplicity here we assume that $Y_i \in \{0, 1\}$. 

The \emph{standard assumption} is that there exist hidden instance labels $y_{ij} \in \{0, 1\}$ that relate to the bag labels as follows: a bag is positive if and only if it contains at least one positive instance. Another way to interpret this assumption is that most positive instance of the bag decides the bag label. Over the years, several other assumptions have been proposed~\citep{foulds2010review}. A common assumption is the \emph{collective assumption}, where all instances (rather than only the most positive one) contribute to the bag label.

Originally, the goal in MIL was to train a bag classifier $f_B$ to label previously unseen bags. Several MIL classifiers do this by inferring an instance classifier $f_I$, and combining the outputs of the bag's instances, for example by the noisy-or rule, $f_B(X_i) = \max_k\{f_I(\mathbf{x}_{ij})\}$. Another group, \emph{bag-level} classifiers, typically represent each bag as a single feature vector and use supervised classifiers for training $f_B$ directly. Such classifiers are often robust, but usually can not provide instance labels. Following \citet{quellec2017multiple}, we refer to methods that can provide instance labels as ``primarily bag level'' and methods that cannot as ``exclusively bag level''. For an in-depth review of MIL (not limited to medical imaging, see~\citep{amores2013multiple,herrera2016multiple,carbonneau2017multiple}.

Because instance-level and some bag-level classifiers can provide instance labels, the focus of MIL became two-fold: classifying bags and classifying instances. This distinction also exists in medical imaging, as discussed in the next section.

\subsection{MIL in medical imaging}

MIL is a natural learning scenario for medical image analysis because labels are often not available at the desired granularity. The goal is therefore to exploit weaker bag labels for training. This idea can be used for different types of tasks. We adopt a categorization similar to that of~\citep{quellec2017multiple}: global detection, i.e. classifying the image as having a target pattern, local detection, i.e. classifying an image patch as having a target pattern, and false positive reduction, i.e. classifying an candidate lesion as true or false positive. \citet{quellec2017multiple} also discuss a ``miscellaneous categorization'' category, however, we find that this is very similar to ``global detection''.

The contributions are summarized in Table \ref{tab:mil}. The most common scenario where MIL is used, is global detection - classifying an entire image as having a particular disease or not. We discuss this in Section~\ref{sec:mil_global}. However, instance classification - local detection - is also relevant. These goals are sometimes pursued simultaneously (Section~\ref{sec:mil_global_local}). If only global detection is addressed, often local detection is relevant, but could not be addressed due to lack of labeled instances. We also briefly discuss local detection only (Section~\ref{sec:mil_local}).

Another application of MIL is false positive reduction, for example classifying candidate lesions that may have been extracted by other algorithms. In this context, the candidate is the bag, and a different viewpoint (such as a different patch) of the candidate is an instance. In other words, the instance has an ``is a'' relationship to the bag, the instances can be highly correlated, and no instances are truly negative. Here the goal is to classify the bag, and instance classification is not as relevant as in global/local detection. We discuss this scenario in Section~\ref{sec:mil_fp}.

\begin{table*}[]
\centering
\footnotesize
\begin{tabular}{llll}

Reference & Application & MIL category & Method\\

\hline
\multicolumn{3}{c}{Brain}\\
\hline

\cite{tong2014multiple} & AD classification & global & excl bag \\ 
\cite{chen2015identification} & cerebral small vessel disease detection & global & instance\\ 
\cite{dubost2017gpunet} & enlarged perivascular space detection  & local & instance \\ 

\hline
\multicolumn{3}{c}{Eye}\\

\hline
\cite{venkatesan2015simpler} & diabetic retinopathy classification & global & excl bag \\ 
\cite{quellec2012multiple} & diabetic retinopathy classification & global, local & instance \\
\cite{schlegl2015predicting} & fluid segmentation & local & instance \\
\cite{manivannan2016subcategory} & retinal nerve fiber layer visibility classification & global, local & instance\\ 
\cite{lu2017multiple} & fluid detection & global & instance \\ 

\hline
\multicolumn{3}{c}{Breast}\\

\hline
\cite{maken2014multiple} & breast cancer detection & global & multiple \\
\cite{rosa2015multiple} & breast cancer detection & global, local & excl bag \\ 
\cite{shin2017joint} & mass localization, classification & global, local & instance\\

\hline
\multicolumn{3}{c}{Lung}\\
\hline

\cite{dundar2007multiple}  & pulmonary embolism detection & false positive & instance \\
\cite{bi2007multiple}& pulmonary embolism detection & false positive & instance  \\
\cite{liang2007computer} & pulmonary embolism detection & false positive & instance \\ 
\cite{cheplygina2014classification} & COPD classification & global & multiple \\
\cite{melendez2014novel} & tuberculosis detection & global, local & instance \\ 
\cite{stainvas2014cancer} & lung cancer lesion classification & false positive & instance \\
\cite{melendez2016combining} & tuberculosis detection & global, local & instance \\ 
\cite{kim2016scale} & tuberculosis detection & global, local & instance \\
\cite{shen2016learning} & lung cancer malignancy prediction & global, local & instance \\
\cite{cheplygina2017transfer} & COPD classification & global & instance \\ 
\cite{li2017thoracic} & abnormality detection (14 classes) & global, local & instance \\ 

\hline
\multicolumn{3}{c}{Abdomen}\\
\hline
\cite{dundar2007multiple}  & polyp detection & false positive & instance \\
\cite{wu2009min} & polyp detection & false positive & instance \\ 
\cite{lu2011effective} & polyp detection, size estimation & false positive & instance \\
\cite{wang2012seeing} & polyp detection & false positive & instance \\ 
\cite{wang2015computer} & lesion detection & global & prim bag\\
\cite{wang2015optimizing} & lesion detection & global & prim bag \\ 

\hline
\multicolumn{3}{c}{Histology/Microscopy}\\
\hline
\cite{dundar2010multiple} & breast lesion detection & global & instance \\ 
\cite{samsudin2010nearest} & pap smear classification & global & multiple \\
\cite{mccann2012automated} & colitis detection & global & instance \\
\cite{zhang2013automated} & skin biopsy annotation & global & multiple \\ 
\cite{kandemir2014empowering} & breast cancer detection & global & excl bag\\
\cite{xu2014weakly} & colon cancer detection & global, local & instance \\
\cite{hou2015efficient} & glioblastoma, low-grade glioma detection & global & instance \\
\cite{li2015multiple} & breast cancer detection & global & prim bag \\ 
\cite{mercan2016multi} & breast cancer detection & global & instance \\
\cite{kraus2016classifying} & cell type classification & global, local & instance \\
\cite{jia2017constrained} & cancerous region segmentation (colon) & global, local & instance \\
\cite{tomczak2017deep} & breast cancer detection & global & instance \\

\hline
\multicolumn{3}{c}{Multiple}\\
\hline
\cite{vural2006batch} & abnormality detection in three applications & false positive & instance\\
\cite{kandemir2015computer} & abnormality detection in two applications & global, local & multiple \\
\cite{hwang2016self} & lesion detection in two applications & global, local & instance \\

\hline
\multicolumn{3}{c}{Other}\\
\hline
\cite{situ2010boosting} & dermoscopic feature annotation & global & prim bag \\ 
\cite{liu2010lesion} & cardiac event detection &  global & instance \\ 
\cite{yan2016multi} & bodypart recognition & global & instance \\ 

\end{tabular}
\caption{Overview of multiple instance learning applications. The third column refers to the type of problem addressed - global and or local detection or false positive reduction. The fourth columns refers to the type of classifier used - exclusively (excl bag) or primarily (prim bag) bag-level, or instance-level.}
\label{tab:mil}
\end{table*}

\normalsize

\subsection{Global detection}\label{sec:mil_global}

The majority of papers on MIL address global detection. The use of MIL is motivated by the fact that strong labels that would enable using supervised learning for local (and therefore also global) detection are not available. Weak labels are available more readily, but may not apply to the entire scan. 

This approach is suitable for many different applications, the more common ones being detection of diabetic retinopathy in retinal images~\citep{venkatesan2015simpler,quellec2012multiple,kandemir2015computer} and detection of cancerous regions in histopathology images~\citep{kandemir2015computer,xu2014weakly,li2015multiple}. Although weak labels are available more easily, the datasets can still be quite small, starting at just 58 bags in a dataset of breast cancer in microarray images~\citep{kandemir2015computer}. Others, such as datasets of COPD in chest CT images~\citep{cheplygina2014classification} or tuberculosis in chest x-ray~\citep{melendez2014novel,melendez2016combining} are in the order of a thousand scans. Only recently, very large datasets started appearing, such as the dataset of 100K chest x-rays used in \citep{li2017thoracic}. 

Global detection can be achieved both with instance-level methods and bag-level methods. Overall, bag-level methods seem to be more successful due to their ability to not treat instances independently (as instance-level methods would), but instead consider the correlations and structure of the instances. In such cases a MIL method can even outperform a fully supervised method~\citep{kandemir2015computer,wang2015computer,vural2006batch,samsudin2010nearest}, showing that the lack of strong labels is not the only use case for MIL. 

In some cases, these scenarios where it is best not to consider instances independently, are not referred to as MIL, but ``batch classification''~\citep{vural2006batch} or ``group-based classification''~\citep{samsudin2010nearest}. An overview of these scenarios and their relationships to MIL can be found in \citep{cheplygina2015on}.

\subsection{Global and local detection}\label{sec:mil_global_local}

Several papers focus both on global and local detection. For example in detection of tuberculosis~\citep{melendez2014novel,melendez2016combining} it is important to both classify the image as having tuberculosis, and highlight the tuberculosis lesions in the image. In fact, in all papers where global detection is the focus, a local detection task could be defined. However, these local detection tasks are often not evaluated, since no labels are available for validation, for example~\citep{cheplygina2015label,kandemir2015computer}.

When both tasks can be addressed, this is done with either instance-level or primarily bag-level method, that can provide instance labels. However, solving two tasks with a single classifier introduces a problem, often overlooked in literature - that the best bag classifier is not necessarily the best instance classifier and vice versa. \citet{cheplygina2015label} demonstrate that the best bag classifier can lead to unstable instance predictions, when trained on bootstrapped versions of the training set. \citet{kandemir2015computer} compare several classifiers on a dataset of Barrett's cancer diagnosis in histopathology image, for which both bag-level and instance-level labels are available. The best bag classifier is an exclusively bag-level method,  while the best instance classifier is an instance-level method that performs reasonably well on bags, but does not have the highest performance. 

Papers where both global and local labels are available for training, show similar results. \citet{li2017thoracic} use both a large number of bag labels and a smaller number of instance labels to train a classifier for global and local detection of various chest x-ray abnormalities. The results show that, when instance classification is the goal, adding more labeled bags does not necessarily increase instance-level performance. \cite{shin2017joint} use both bag and instance labels for localization and classification of breast masses. They show that bag labels should be given less weight than the instance labels - i.e. using all the labels together does not lead to the best results. 

An important aspect of classifiers doing both global and local detection, is their explanation of the global label in terms of local labels, for example, highlighting abnormalities in an image. If the classifier is trained with global labels, it could happen that it only detects the most abnormal part of an image where multiple abnormalities are present. This creates an issue for the interpretability of the method. Currently work on attention mechanisms such as~\cite{ilse2018attention} is investigating solutions to this problem.

\subsection{Local detection only?}\label{sec:mil_local}

Given that we have covered global detection, and a combination of global and local detection, it would seem that local detection is the next logical category. Indeed, recently methods that focus only on the local detection have emerged. However, in such cases global detection is still a task that is being optimized for, so these papers can also be seen as falling under the ``global and local detection'' category. 

Why have a ``local detection only'' category, if the category is technically empty? We decided to retain this section to explicitly address what might be a perceived difference. Methods focusing on local detection are often referred to as ``weakly supervised''. This term is sometimes interchangeably used with MIL, but seems to be common when global detection is not addressed. This might create a false impression that MIL and weak supervision (weak referring to only having bag labels) are disjoint, which is not the case.

On the other hand, not all papers that call themselves weakly supervised, fall under the MIL category. For example, \citet{donner2009weakly} uses ``weakly supervised'' to refer to ``few annotations'' that are used to initialize label propagation. In our classification this would be a SSL method. \citet{rajchl2016learning} uses ``weakly supervised'' to describe that the labels are noisy, which is a different variation not covered in this survey.

\subsection{False positive reduction with multiple viewpoints} \label{sec:mil_fp}

A task that also focuses on classification of bags, but uses different assumptions, is known as false positive reduction within MIL~\cite{quellec2017multiple}. Here the bag represents a candidate tumor or lesion (possibly detected by a different method), and the instances represent different viewpoints~\citep{wu2009min,lu2011effective,wang2012seeing}. The bag label in principle applies to all the instances, since they are versions of the same candidate, which is different from the other MIL scenarios with label ambiguity. However, similar to global detection, a MIL classifier can outperform a supervised classifier, because it benefits from combining information of different instances, to classify the bag. 

A difference from global detection is that the instance classification task is less relevant. Since the goal is to classify the candidate as a whole, and the assumption is that all instances have the same label, it might be less interesting to find out which instances contributed the most to the candidate's label.

\section{Transfer learning}\label{sec:tl}

Another popular learning scenario is transfer learning~\citep{pan2010survey}. Here the goal is to learn from related learning problems. One example of related, but different learning problems. One example is due to differences between acquisition of images, such as the use of different scanners or scanning protocols. Another example is related classification tasks for the same data, such as detection of different types of abnormalities.

More formally, in the scenarios covered so far we assumed that the training and test data are from the same \emph{domain} $\mathcal{D} = (\mathcal{X}, p(X)$), defined by the feature space and distribution of the samples. We also assumed they addressed the same \emph{task} $\mathcal{Y} = (\mathcal{Y}, f(\cdot))$ defined by the label space and the mapping between the feature and the label space. In transfer learning scenarios, we assume that we are dealing either with different domains $\mathcal{D}_S \neq \mathcal{D}_T$ and/or different tasks $\mathcal{T}_S \neq \mathcal{T}_T$.

For example, such differences can be caused by different marginal distributions $p(\mathbf{x})$, different labeling functions $p(y|\mathbf{x})$, or even different feature and label spaces. In our illustrative example in Figure\ref{fig:learning_scenarios}, the $\mathbf{x}$'s are the feature vectors describing the appearance of lung ROIs, and the $y$'s are the categories the patches belong to. Changes in subject groups, scanners and scanning protocols, can affect the distributions $p(\mathbf{x})$, such as ``this dataset has lower intensities'', $p(y)$, such as ``this dataset has a large proportion of emphysema'' and/or $p(y|\mathbf{x})$, such as ``in this dataset this appearance corresponds to a different category''. One or more of these differences mean that the distribution $p(D_S)$ of the training or source set is different from the distribution $p(D_T)$ of the test or target set. 

Transfer learning approaches addressing these scenarios can be grouped by what they transfer. In this survey we focus on instance transfer i.e. assuming that source data can be reweighted to train the target classifier, and feature transfer i.e. encoding knowledge from the source domain into the feature representation for the target domain. 



\subsection{TL in medical imaging}

The papers using TL are summarized in Table \ref{tab:tl}. We discuss these methods based on whether the tasks or the domains are different, or both. 

In the same domain, different task scenario (Section~\ref{sec:tl_same_different}), we are often dealing with multiple tasks for the same set of images, such as detecting multiple types of abnormalities, where detection of each type of abnormality is a binary classification problem. Although the label spaces $\mathcal{Y}$ are the same for each task, the labeling functions $f$ are different, leading to $\mathcal{T}_S \neq \mathcal{T}_T$.

This scenario is often approached with feature transfer - learning features that are relevant for multiple tasks, thus effectively increasing the sample size and/or regularizing the classifier. An in-depth explanation of why this works can be found in~\cite{ruder2017overview}. Rather than training multiple tasks simultaneously, representation learning approaches where an (unsupervised) task such as reconstructing the data, is done first, are also possible.

In the ``different domain, same task'' scenario (Section~\ref{sec:tl_different_same}), we are dealing with, for example, data acquired with different scanners. This can cause differences in the distributions of the samples $p(\mathbf{x})$ leading to differences in domains $\mathcal{D}_S \neq \mathcal{D}_T$. It is also possible that there are differences in the labeling functions. This means the source and target tasks are not strictly the same, but this can still be assumed by the method. 

This scenario is often addressed with instance transfer. Instance transfer involves, for example, weighting source training samples such that only relevant samples receive high weights, or realigning the source domain with the target domain with the goal of bringing $p(\mathbf{x}_S)$ and $p(\mathbf{x}_T)$ closer together. After this, the union of the weighted instances can be used for training. The alignment approach can be referred to as feature transfer by others, because the features are being adapted. However, we group these approaches together since they are all aimed at decreasing the number of irrelevant samples, and/or increasing the number of relevant samples.

Finally, there is also a different task, different domain scenario (Section~\ref{sec:tl_different_different}). Although according to \citep{pan2010survey} this would fall under ``unsupervised transfer learning'' and only address clustering, we find that this is also relevant in the supervised case, through feature transfer. In this case, the source task is used to pretrain a network. The network can then be used in two strategies~\citep{litjens2017survey}: for feature extraction, or as a starting point for further training (fine-tuning) of the target task. Both are currently very popular in medical image analysis.

\begin{table*}[]
\centering
\scriptsize
\begin{tabular}{l p{4cm} l l l}

Reference & Topic & Task  & Domain & Transfer type\\
\hline
\multicolumn{5}{c}{Brain}\\
\hline

\cite{zhang2012multi} & MCI conversion prediction & different & same & feature, multi-task \\

\cite{wang2013modeling} & tissue, lesion segmentation & same & different & instance, weight\\ 

\cite{opbroek2014transfer} & tissue, lesion segmentation  & same & different & instance, weight\\
\cite{guerrero2014manifold} & AD classification & same & different & instance, align \\

\cite{opbroek2015weighting} & tissue, lesion segmentation  & same & different & instance, weight\\

\cite{cheng2015multimodal} & MCI conversion prediction & different & same & feature,  multi-task \\

\cite{goetz2016dalsa} & tumor segmentation  & same & different  & instance, weight \\ 

\cite{wachinger2016domain} & AD classification  & same & different  &  instance, weight \\

\cite{cheplygina2016asymmetric} & tissue segmentation & same & different & instance, weight \\ 

\cite{ghafoorian2017transfer} & lesion segmentation  & same & different & feature, pretraining \\

\cite{kamnitsas2017unsupervised} & segmentation of abnormalities & same & different & feature, pretraining \\ 

\cite{alex2017semisupervised} & lesion segmentation  & different & same & feature, pretraining \\

\cite{hofer2017simple} & AD classification  & same & different & instance, align\\

\cite{hon2017towards} & AD classification  & different & different & feature, pretraining \\

\cite{kouw2017mr} & tissue segmentation & same, different & instance, align\\


\hline
\multicolumn{5}{c}{Breast}\\
\hline
\cite{huynh2016digital} & tumor detection  & different & different & feature, pretraining \\
\cite{samala2016mass} & mass detection  & same & different & feature, pretraining \\

\cite{kisilev2016medical} & lesion detection, description in mammography or ultrasound & different & same & feature, multi-task \\ 

\cite{huynh2017comparison} & chemotherapy response prediction & different & different & feature, pretraining \\

\cite{dhungel2017deep} & mass detection, classification & different & same & feature, pretraining \\

\hline
\multicolumn{5}{c}{Lung}\\
\hline

\cite{bi2008improved} & abnormality classification  & different & same & feature,  multi-task \\ 

\cite{schlegl2014unsupervised} & lung tissue classification  & different & same/different & feature, pretraining\\ 

\cite{bar2015chest} & chest pathology detection & different & different & feature, pretraining\\

\cite{ciompi2015automatic} & nodule classification & different & different & feature, pretraining \\

\cite{shen2016learning} & lung cancer malignancy prediction & different & same & feature,  multi-task \\ 

\cite{chen2017automatic} & attribute classification in nodules & different & same & feature, multi-task \\

\cite{hussein2017risk} & attribute regression, malignancy prediction & different & same & feature, multi-task \\

\cite{cheplygina2017transfer} & COPD classification & same & different & instance, weight\\

\cite{christodoulidis2017multisource} & ILD classification & different & different/same & feature, pretraining/multi-task \\

\hline
\multicolumn{5}{c}{Abdomen}\\
\hline

\cite{ravishankar2016understanding} & kidney detection & different & different & feature, pretraining \\ 
\cite{nappi2016deep} & polyp detection & different & different &  feature, pretraining  \\ 
\cite{sonoyama2016transfer} & endoscopic image classification & same & different & instance, align \\
\cite{azizi2017transfer} & prostate cancer detection & same & different & feature, pretraining\\
\cite{cha2017bladder} &  bladder cancer treatment response prediction & different & different & feature, pretraining\\
\cite{chen2017transfer} & prostate cancer classification & different & different & feature, pretraining\\
\cite{meng2017liver} & liver fibrosis classification  & different & different & feature, pretraining \\ 
\cite{li2017exploring} & gastrointestinal bleed detection& different & different & feature, pretraining\\ 
\cite{ribeiro2017exploring} & polyp classification in endoscopy & different & same/different & feature, pretraining\\ 
\cite{mahmood2017unsupervised} & depth estimation in endoscopy & same & different & feature, pretraining  \\ 
\cite{ross2017exploiting} & surgical instrument segmentation & same & different & feature, pretraining\\ 
\cite{zhang2017automatic} & polyp detection & different & different & feature, pretraining \\

\hline
\multicolumn{5}{c}{Histology and microscopy}\\
\hline

\cite{ablavsky2012transfer} & mitochondria segmentation  & same & different & instance, regularization \\ 
\cite{becker2014domain} & mitochondria segmentation & same & different & instance, align\\
\cite{kandemir2015asymmetric} & tumor detection & same & different & instance, align\\
\cite{bermudez2016scalable} & organelle segmentation & same & different & instance, regularization  \\ 
\cite{gadermayr2016domain} & glomeruli detection & same & different & instance, weight\\ 
\cite{chang2017unsupervised} & tissue classification & different & same & feature, pretraining \\
\cite{phan2016transfer} & staining pattern detection & different & different & feature, pretraining \\
\cite{murthy2017center} & visual attribute classification & different & same & feature,  multi-task\\
\cite{huang2017epithelium} & epithelium stroma classification & same & different & feature, pretraining \\
\cite{spanhol2017deep} & breast cancer classification & different & different & feature, pretraining\\

\hline
\multicolumn{5}{c}{Multiple}\\
\hline

\cite{hwang2016self} & lesion detection, 2 applications  & different & same & feature, multi-task \\

\cite{moeskops2016deep} & segmentation, 3 applications & different & different & feature,  multi-task\\

\cite{tajbakhsh2016convolutional} & detection and segmentation, 4 applications & different & different & feature, pretraining\\

\hline
\multicolumn{5}{c}{Other}\\
\hline

\cite{bi2008improved} & heart segment classification & different & same & feature,  multi-task \\

\cite{ciompi2010fusing} & plaque classification & same & different & instance, weight \\

\cite{heimann2014real} & US transducer localization & same & different & instance, weight \\
\cite{chen2015standard} & US standard plane localization & different & different & feature, pretraining\\ 

\cite{engelen2015multi} & carotid plaque component segmentation
 & same & different & instance, weight \\

\cite{antony2016quantifying} & osteoarthritis quantification & different & different & feature, pretraining \\ 

\cite{conjeti2016supervised} & tissue classification & same & different & instance, align \\ 
\cite{elmahdy2017low} & skin lesion classification & different & different & feature, pretraining \\ 

\cite{murphree2017transfer} & melanoma classification & different & different & feature, pretraining \\ 
\cite{liu2017feature} & thyroid nodule classification  & different & different & feature, pretraining \\ 
\cite{menegola2017knowledge} & melanoma classification & different & different & feature, pretraining \\ 

\end{tabular}
\caption{Overview of transfer learning applications. The last column refers to the type of transfer approach, i.e. whether it is instance transfer (by weighting or aligning samples) or feature transfer (by pretraining on an auxiliary task in the same or different domain, or multi-task learning)}.
\label{tab:tl}
\end{table*}

\normalsize

\subsection{Same domain, different tasks} \label{sec:tl_same_different}

Perhaps the earliest way in which transfer of information was leveraged within medical imaging, is inductive transfer learning, or learning different tasks within the same domain. For example, in lung images, we might be interested in detecting different types of abnormalities. Rather than learning a multi-class task, or learning several binary tasks independently, we could learn these binary tasks jointly. The intuition is that these tasks will share task-independent features, and learning these tasks jointly increases the amount of data, leading to a more robust representation. This scenario includes \emph{multi-task learning} (MTL), where a lot labeled source data is available, and self-taught learning, where no labeled source data is available.

We find that in medical imaging, many works fall under the multi-task learning scenario. ~\cite{bi2008improved} describe a probabilistic framework for MTL algorithms and apply it to two applications with different characteristics. The first application is classifying nodules in chest CT, while also using labeled examples of ground glass opacities. Even though the tasks have different label spaces and the datasets of nodules and ground glass opacities cannot be directly combined, learning the tasks jointly increases the effective sample size. 

The second application is classifying multiple heart wall segments per subject. Instead of classifying each segment independently, they simultaneously classify all segments, essentially predicting a vector of labels per subject. This does not increase the sample size, but still benefits the classifier through regularization. The authors also demonstrate that MTL has the largest advantage at low sample sizes, where regularization is most needed.

Multi-task learning is also used in classification of Alzheimer's disease in brain MR images. Usually subjects are classified into AD, MCI and cognitively normal (CN) classes. Additionally, MCI subjects can be classified into converters (to AD) and non-converters. Here again there are two main strategies. Effective increase of the sample size can be seen in \citep{cheng2015multimodal} for MCI conversion prediction, where these tasks can be combined even though the label spaces are different. 

Using multiple labels for the same samples can be seen in \citep{zhang2012multi}. Motivated by the fact that the underlying pathology influences both the diagnosis (Alzheimer's, mild cognitive impairment or cognitively normal) and two cognitive scores, they predict these three labels simultaneously. In a further experiment, they predict the change in these labels, i.e. the absolute change in the cognitive scores, and whether the MCI subjects convert to AD or not. 

Other applications where multiple labels are predicted include classification of lung diseases~\citep{li2017thoracic}, and classification of visual attributes of images, such as attributes of lung nodules\citep{chen2017automatic,hussein2017risk} or skin lesions~\citep{murthy2017center}.

Finally, there are a couple of examples of self-taught learning, where there are labels for only one of the tasks. This happens in scenarios where one dataset needs to address multiple tasks, for example localization of abnormalities and their classification~\citep{hwang2016self}, or description~\citep{kisilev2016medical}. There are then two optimization problems being solved using the same labels. Note that while \citeauthor{pan2010survey} call these works ``self-taught learning'', in practice other names may be used, such as ``self-transfer learning'' \citep{hwang2016self} or multi-task learning \citep{kisilev2016medical}. There is a relationship between these works and MIL, which we will explore in the discussion. 

In the examples above, multi-task learning is done by sharing the weights or parameters for the model, but using different outputs depending on the task. For example, in deep learning, this could be achieved by sharing the hidden layers, but using a different set of output layers. The label space for each of the tasks is therefore the same, as if that task was learned individually. An exception is \cite{moeskops2016deep}, where multiple tasks - tissue segmentation in MR, pectoral muscle segmentation in MR and coronary artery segmentation in CT - are learned in a joint label space, like a multi-class problem. While in principle this means that confusion between tasks could occur, for example a voxel of brain tissue could be classified as a voxel of the coronary artery, the results show that most errors happen within the same task.

Another way to use different tasks within the same domain, is by learning the tasks sequentially, rather than in parallel, as in multi-task learning. For example, \citet{dhungel2017deep} first train a regression model to predict handcrafted features that are known to be related to the target labels.  This model is then used for initialization of the target model. Although the handcrafted features are used as labels, they are not provided by experts, so this type of pretraining can be considered to be unsupervised.

There are other ways to add such unsupervised tasks to improve the target (supervised) task. An approach that is gaining popularity is finding a representation from which (part of) the data can be reconstructed. For example, \citet{ross2017exploiting} first decolorize their training images, then use recolorization as an additional task to learn a good representation.

Related to this idea is adversarial machine learning~\citep{biggio2010adversarial}, with generative adversarial networks (GANs)~\citep{goodfellow2014generative} as a popular technique. GANs work by having an interplay between two networks - a generator, that generates samples based on the training set distribution, and a discriminator, that classifies such samples as either real or generated. By competing with each other, the networks learn a good representation of the data. In a sense, this is an example of learning from multiple tasks on the same data.


\subsection{Different domains, same task} \label{sec:tl_different_same}

Other early efforts in transfer learning in medical imaging focus on the scenario where the classification task is the same, but the domains are different, for example due to the use of data from different hospitals. Due to the differences in data distributions, it may not be optimal to simply train a classifier on one domain, and then test it on the other domain, or to train a classifier on the union of all available labeled data. 

Changes in distributions can occur due to several reasons. For example, \cite{opbroek2015weighting,opbroek2014transfer,kouw2017mr} address segmentation of MR data from different scanners, which alters the appearance of the images, and different populations, which changes the distribution of classes in the data. \cite{ciompi2010fusing,conjeti2016supervised} address differences between in vitro and in vivo ultrasound, where the absence/presence of blood flow causes a distribution shift. \cite{bermudez2016scalable} focus on segmentation of cells in microscopy images of different parts of the brain, which results in heterogeneous appearances.

The methods that address these distributions are mainly instance-transfer methods. One strategy is to change the source distribution by weighting the instances for training, such that the source distribution matches the target distribution as closely as possible. This is possible via importance weighting, where each instance is assigned a weight based on probability of belonging to the target domain. This strategy is optimal if only the marginal distributions are different but the labeling functions are the same, but in practice can also be helpful with different labeling functions~\citep{opbroek2015weighting,cheplygina2017transfer}. Weights can also be assigned on other characteristics, without explicitly addressing the distributions of the feature vectors. When classifying subjects as having Alzheimer's, \cite{wachinger2016domain} perform weighting based on patient characteristics such as age, while these factors are not used by the classifier.

Another instance-transfer strategy is to align the source and target domains by a transformation of the feature space. Once the domains are aligned, the instances of the source domain can be used for training. \cite{conjeti2016supervised} use principal component analysis to align in vitro and in vivo ultrasound images in feature space as a preprocessing step, before training a random forest on the source data and adapting it with the (aligned) target data. \cite{guerrero2014manifold} align subjects from 1.5 Tesla and 3 Tesla scanners by exploiting correspondences between the two domains. A correspondence-free approach to align representations of MR scans from different datasets is used by \cite{hofer2017simple}, by assuming a Gaussian distribution for each dataset. \cite{kouw2017mr} use pairs of similar (same class) and dissimilar (different class) voxels from scans acquired with different scanners to learn an invariant feature representation. Training on the union of the voxels using this representation outperforms training on source or target data only, or the union of the voxels with the original representation. 

Another difference between methods is whether they assume the presence of labeled data from the target domain.  Unsupervised transfer is addressed in \citep{wang2013modeling,heimann2014real,cheplygina2017transfer,opbroek2015weighting} among others. Other works such as ~\citep{conjeti2016supervised,wachinger2016domain,goetz2016dalsa,opbroek2014transfer} focus on supervised transfer, with a small amount of labeled data from the target domain.

\subsection{Different task, different domains}
\label{sec:tl_different_different}

With the development of deep learning methods, it has become more common to transfer information between different tasks and different domains. The idea behind this is to find a good feature representation. This is achieved when a lot of source data is available, which can be used to train a deep network. This pretrained network can then be used to extract ``off-the-shelf'' features from the target dataset, or as a starting point for further training or fine-tuning the network to the target task. Such strategies are often compared to training a network ``from scratch'', i.e. without using transfer.

The source data can be from a totally different task. Using non-medical images as source data is now common for 2D networks. Probably the first work to do this is \cite{schlegl2014unsupervised}. For the target task of classifying tissue types in chest CT slices, they used three different source tasks: natural images, other chest CT images, and head CT images. They found that natural images performed comparably or even slightly better than using only lung images. Using brain images was less effective, possibly due to large homogeneous areas present in brain CT, but not in lung CT, which has more texture information. After this work, more results showing transfer from non-medical images appeared, for example~\citep{bar2015chest,ciompi2015automatic}.

Transfer from natural images is used often in practice. Common datasets used for transfer are datasets annually released by the Imagenet Large Scale Visual Recognition Challenge~\citep{russakovsky2015imagenet}. The datasets have more than a million images and thousand categories of everyday objects. Since this methodology is so popular, we are not able to provide an exhaustive list of papers that apply it, and focus on papers that investigate underlying causes of when transfer is successful or not.

For detecting and classifying colorectal polyps, \cite{zhang2017automatic} transfer from Imagenet (1.2 million images in 1000 categories) and from Places, a scene recognition dataset of 2.5 million images in 205 categories such as ``bathroom''~\citep{zhou2017places}. \cite{zhang2017automatic} hypothesize that Places has higher similarity between classes than Imagenet, which would help distinguish small differences in polyps. This indeed leads to higher recognition rates, also while varying other parameters of the classifier.

\cite{menegola2017knowledge} compare off-the-shelf features and finetuning strategies for the task of melanoma classification, and use two datasets for pretraining: Imagenet and Kaggle Diabetic Retinopathy (KaggleDR) with 35K images\footnote{https://www.kaggle.com/c/diabetic-retinopathy-detection/data}. KaggleDR contains retinal images that are in a sense similar to melanoma images, capturing a single object of interest. The authors find that finetuning outperforms the off-the-shelf strategy, and that transfer from Imagenet only is more successful than transfer from KaggleDR, or from the union of the datasets. Although the advantage of Imagenet over KaggleDR could perhaps be explained by the dataset size, the fact that the union of the datasets performs worse, indicates that there are more factors to be considered.

\cite{ribeiro2017exploring} investigate pretraining and fine-tuning of different source datasets for classification of polyps in endoscopy images. Different from the previous papers, they extract datasets of the same number of classes and images from the available types of data, making it a more fair comparison. They find that texture datasets perform best as source data, outperforming other datasets of endoscopy images. They also note that increasing the number of images and classes does not always improve performance.

These results do not always hold. In a study of predicting response to cancer treatment in the bladder, \cite{cha2017bladder} compare networks without TL, networks pretrained on natural images, and networks pretrained on bladder ROIs. They find that there are no statistically significant differences between the methods.

These results are interesting if we consider that more traditional transfer learning methods focused on increasing the similarity between the source and target data. The results summarized here suggest that there is a trade-off between similarity, size and perhaps diversity of the source data.

\section{Discussion}\label{sec:discussion}

\subsection{Trends}

\begin{figure}
    \centering
    \includegraphics[width=0.9\columnwidth]{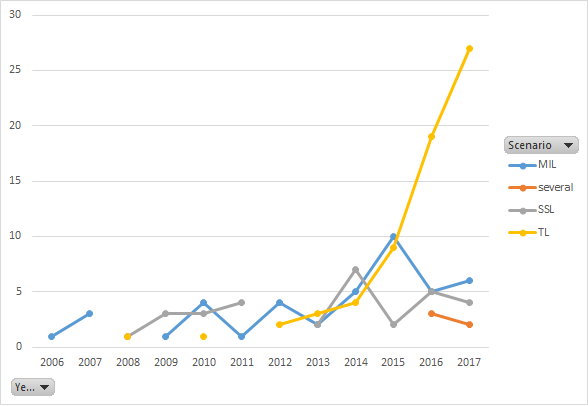}
    \caption{Number of discussed papers by year, grouped by learning scenario.}
    \label{fig:scenario_year}
\end{figure}

We first examine the overall trends in the use of different learning scenarios. Fig.~\ref{fig:scenario_year} shows how the surveyed papers for each scenario are distributed across different years. Transfer learning is clearly the most popular, although this has only become evident in recent years. A reason for this might be the availability of datasets and tools. For SSL and MIL, a specific type of data/labels need to be available, while for TL, it is possible to use a completely external dataset in addition to the target data, and pretrained models can be easily downloaded. 




\begin{figure}
    \centering
    \includegraphics[width=0.9\columnwidth]{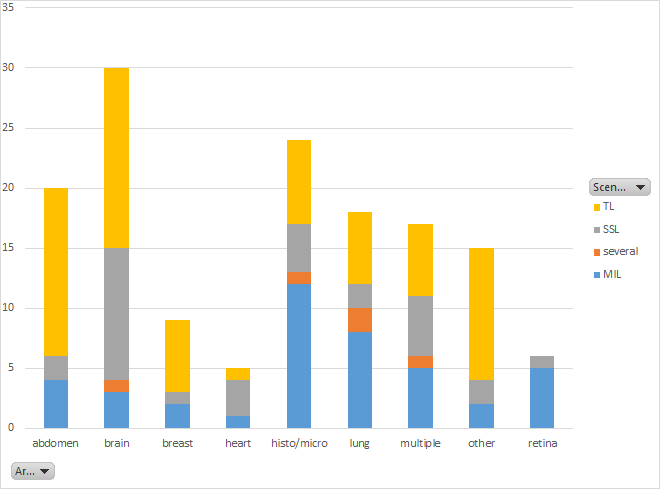}
    \caption{Distribution of papers across learning scenarios and applications.}
    \label{fig:scenario_application}
\end{figure}

There are also trends related to the different application areas. In this paper we have used the following categories, inspired by \citet{litjens2017survey}: brain, retina, chest, breast, heart, abdomen, histology/microscopy and other applications. Fig.~\ref{fig:scenario_application} shows the distribution of these applications across the learning scenarios. Overall, brain is the most common application, followed by histology/microscopy and the abdomen. Breast, heart and retina, on the other hand, have relatively few papers. Around 10\% of the papers address multiple applications.

The application also influences the popularity of different learning scenarios. For example, MIL is frequently used for histology/microscopy, but is not as common for tasks within the brain.  One reason is that in histology/microscopy it is more reasonable to assume that the patches within an image do not have an ordering, and there can be a variable number of patches per image, as is the case in multiple instance learning. However, this is less suitable for the brain, where anatomical correspondences can be determined and are informative, and the MIL scenario is less applicable.

\subsection{Related learning scenarios}

A gap in the current literature is that relevant learning scenarios are sometimes not considered. However, doing so could further our understanding of the underlying classification problem, possibly leading to a better fit between the problem, assumptions and the method used.

One example of relevant learning scenarios is SSL and MIL. MIL can be seen as a special case of SSL~\citep{zhou2007relation} if the traditional assumption is used, because the instances in negative bags can be considered labeled, and the instances in positive bags can be considered unlabeled, but with additional constraints that not all instances in a positive bag can be negative. Investigating this relationship by comparing methods with and without such constraints, could help elucidate the importance of the bag structure.

SSL and TL are also related. When dealing with a set of labeled data and a set of unlabeled data for the same task, without knowledge about the domains it could be logical to use a SSL method. On the other hand, if information about domains is available, a TL method would be more appropriate. Comparing the two could help with understanding the differences between datasets - is there perhaps a transfer learning situation where we didn't suspect one before? Since transfer learning has started becoming more popular only recently, it is possible that earlier papers that use multi-center data in SSL, such as \citep{dundar2007multiple}, do not address this issue.

There are several links between MIL and TL. Using MIL can avoid the need to use a TL method, because MIL labels from the same domain can be acquired more easily. This is illustrated in \citep{melendez2014novel}, where instance labels are available only for one domain, but bag labels are available for multiple domains. A MIL classifier trained on same-domain bag labels outperforms a fully supervised classifier trained on different-domain instance labels. It would have been valuable to see how a TL approach would compare to the same-domain MIL method, and to the combination of both.

Another link between MIL and TL is in scenarios where two related classification tasks are addressed, such as global detection and local detection. In MIL methods this is usually achieved by training a single classifier, but we can also view this as an example of multi-task or self-transfer learning, where two classifiers are trained with a shared representation. Finally, we could consider situations where the differences in distributions are only on instance level, or only on bag level, but not both. 

There are also opportunities in exploring related learning scenarios that are not yet common in medical imaging. Positive and unlabeled learning~\citep{elkan2008learning} has received quite some attention in the machine learning community. The idea is to learn from only positive and unlabeled examples, which may happen when the expert misses some positives during annotation. The absence of a positive label does not imply that an example is negative, and thus a non-positive example is considered unlabeled. Different from novelty detection, where only positive examples are used for modeling healthy or normal examples, positive and unlabeled learning aims to still use the unlabeled examples. Although this scenario seems very suitable for medical imaging, the only paper we have found directly addressing this scenario is ~\cite{zuluaga2011learning}. 

Other possibilities include the ``siblings'' of MIL, such as batch classification~\citep{vural2006batch} and group-based learning~\citep{samsudin2010nearest}. We have grouped these works in the MIL section, as they can be seen as variations of MIL with different assumptions~\citep{cheplygina2015on}. Although from the point of literature search it is counterproductive to use such different names for these scenarios, their similarities and differences could help us better understand the diversity of MIL problems being addressed in the literature.

\subsection{Full potential of available data}

The available (labeled) data is not always used to its full potential, possibly due to the constraints of a particular method. For example, papers on MIL may (unnecessarily) convert a regression or multi-class problem into a binary problem, because this is how MIL was traditionally formulated. As a result, different grades or types of a disease can be aggregated into ``healthy'' and ``abnormal''. Others may remove more difficult classes. However, this is not necessary helpful for machine learning methods. For example, ~\cite{menegola2016towards} demonstrate that removing one of the two disease classes results in lower performance, possibly because the method has fewer samples in total to learn from.

An opportunity is to use multiple labels when the ground truth is determined by consensus of different experts. Often the individual labels of the experts are combined into consensus labels, which are then used for training.  However, as  \citet{warfield2004simultaneous} and \citet{guan2017who} show, modeling the individual labelers during training can outperform averaging the labelers in advance.

Another opportunity is using clinical variables as additional outputs for the model. These are currently not used very often, but can improve prediction~\citep{zhou2013modeling}. Even age or sex could be included as additional labels to predict. While not interesting prediction tasks by themselves, these could be leveraged via multi-task learning, for example, by using these as auxiliary tasks or ``hints''~\citep{ruder2017overview}. Clinical reports with more detailed information, such as describing the location of abnormalities, can also provide additional information~\citep{schlegl2015predicting}.

Finally, the data itself can be used for pretraining in an unsupervised manner, for example by reconstructing the data while learning a good representation. This is already being done by a few papers discussed in Section~\ref{sec:tl_same_different}. However, this approach could be an opportunity for other applications where additional images and/or modalities are available, and that could be used as auxiliary tasks.

\subsection{Acquiring additional labels}

While the methods in this survey can certainly improve the robustness of classifiers, we feel there is a limit on what can be achieved without additional labels. Active learning methods such as \citep{melendez2016combining,su2016interactive} aim to minimize the number of labels needed for the same or better performance, by only querying the labels that are most ambiguous or will lead to most improvement for the classifier. Given the same budget for labels, this could potentially lead to better performance overall.

Following the success of crowdsourcing in computer vision, crowdsourcing is also gaining an important place in medical imaging. These methods aim to collect (possibly noisy) labels from the public. When combining multiple annotators, the noise is expected to be reduced. Most studies to date investigated the quality of such labels compared to expert labels~\citep{maier2015crowdtruth,cheplygina2016early,mitry2015crowdsourcing}. Methods that use the crowdsourced labels inside machine learning methods, are less common, for example~\citep{albarqouni2016aggnet}. We expect that this will be an important direction for future research.

\subsection{Generalization}

A main challenge with not-so-supervised learning in medical imaging is that most works are proofs of concept on one or (less frequently) few applications. This makes it difficult to generalize the results and gain insight into whether the method would work in a different problem.

One partial solution would be to vary the characteristics of a single dataset - for example, subsample the training data to create learning curves, change the class priors to investigate the influence of class imbalance, or select or merge different classes. Another partial solution would be to perform ablation experiments, i.e. removing a part of the method's functionality, to understand what factors contribute most to the result.

A related challenge of not generalizing to other applications is publication bias: negative results, and/or results from an existing method may not be published, or published in a less popular venue. \cite{borji2018negative} provides an excellent discussion on why this is detrimental to research in computer vision. We feel that this is something that should also be discussed within the medical imaging community. 

Challenges such as grand-challenges.org are a great resource for benchmarking algorithms on open datasets. However, these too often address only a single application, with the risk of overfitting to these datasets as a community. We see a promising research direction in platforms where the same methods could be applied to a range of datasets from different medical applications.



\section{Conclusion}

We have discussed over 140 papers in medical image analysis that focus on classification in a ``not-so-supervised'' learning scenario, often due to lack of representative annotated data. We focused on semi-supervised, multi-instance and transfer learning, of which transfer learning is the most popular in recent years. While individual papers demonstrate the usefulness of such approaches, there are still many questions on how to best use these methods. We expect future research to benefit from examining the connections between learning scenarios and generalizing the results between applications.

\section*{Acknowledgments}

We thank the following people for their constructive comments on the preprint: Andreas Eklund, Caroline Petitjean, Gwenol\'e Quellec, Jakub Tomczak, Jesse Krijthe, Marco Loog, Maximilian Ilse, Ragav Venkatesan and Wouter Kouw. This work was partly funded by the Netherlands Organization for Scientific Research (NWO), grant no. 639.022.010.

\biboptions{authoryear}
\bibliographystyle{model2-names}
\bibliography{refs_veronika}

\end{document}